\documentclass[letterpaper]{article} 
\usepackage{aaai2026}  
\usepackage{times}  
\usepackage{helvet}  
\usepackage{courier}  
\usepackage[hyphens]{url}  
\usepackage{graphicx} 
\usepackage{natbib}  
\usepackage{caption} 
\usepackage{algorithm}
\usepackage{algorithmic}
\nocopyright
\usepackage{newfloat}
\usepackage{listings}
\DeclareCaptionStyle{ruled}{labelfont=normalfont,labelsep=colon,strut=off} 
\floatstyle{ruled}
\newfloat{listing}{tb}{lst}{}
\floatname{listing}{Listing}
\title{Who Brought Easter Eggs to Eid? Auditing LLM-Generated Cultural Translation of Math Word Problems Across Languages and Regions}

\author{
    Parisa Suchdev\textsuperscript{\rm 1,2,3},
    Juniper Lovato\textsuperscript{\rm 1,2,3,4}
}

\affiliations{
    \textsuperscript{\rm 1}Department of Computer Science, University of Vermont\\
    \textsuperscript{\rm 2}Vermont Complex Systems Institute, University of Vermont\\
    \textsuperscript{\rm 3}Computational Ethics Lab, , University of Vermont\\
    \textsuperscript{\rm 4}Complexity Science Hub\\
    psuchdev@uvm.edu
}

\begin{document}

\maketitle

\begin{abstract}
Large language models are increasingly used to adapt math word problems for personalized learning at scale, but it remains an open question whether those adaptations are consistent across models, preserve cultural diversity at scale, and reveal which cultural entities models treat as most salient. We analyze how Claude Opus 4, GPT 4.1, and Gemini 2.5 Pro adapt 60 English math word problems into Bengali, Hindi, Punjabi (India), Urdu, Sindhi (Pakistan), Italian, and Sicilian (Italy), a language set spanning the full resource spectrum, from high-resource Italian and Hindi to under-studied Sindhi, Sicilian, and Punjabi. We annotate 6,489 entity transformations, coding whether models preserve, localize, generalize, omit, or change entities such as names, foods, and places. Models agree on transformation type in 62.5\% of cases and on specific substitutions in only 33.5\%, meaning model choice directly shapes which cultural world students encounter. All 21 language-model combinations show entropy collapse, with adaptation compressing rather than expanding cultural diversity. Models prioritize surface markers such as names, foods, and currencies while preserving deeper structural features such as grade-level systems that embed culturally specific assumptions. Despite prompts specifying target countries, models misattribute regional context by using Bangladeshi taka for Indian Bengali students and produce cross-cultural contamination, such as adapting egg hunts as Eid activities. Some failures are visible in individual translations. Others, including diversity collapse, systematic preference for surface markers, and consistent regional misattribution, emerge only through corpus-level analysis. The surface plausibility that makes adapted problems look correct is precisely what makes deeper failures easy to overlook.
\end{abstract}

\begin{links}
    \link{Code and Data}{https://github.com/Parisasuchdev/cultural-entity-audit-of-word-problems}
\end{links}

\section{Introduction}
Localizing a math word problem involves more than changing names, currencies, or places. In culturally responsive pedagogy, teachers ground instruction in students' lived experiences, linguistic familiarity, and social worlds \citep{ladson1995toward, gay2018culturally, paris2012culturally}. When a problem assumes unfamiliar settings, practices, names, or currencies, students face an added cognitive burden unrelated to the mathematics \citep{kintsch1985understanding, sweller1988cognitive}, and may feel that the task was written for someone else \citep{walton2007questionbelonging}. Careful adaptation replaces unfamiliar referents with locally meaningful ones while preserving the mathematical structure and difficulty of the task. Each substitution is therefore a pedagogical judgment, shaped by knowledge of the community and of what matters for student understanding \citep{moll2006funds}.

This work is highly skilled and important, and it is increasingly being supported by large language models (LLMs). Educators use LLMs to translate, draft, or adapt math word problems \citep{walkington2025implications}. At the level of a single example, the output can look appropriate. A word problem about dollars and supermarkets, rendered into Sicilian, becomes one about euros and supermercati. The names change, and the currency changes. A student in Palermo receives a problem that appears, on the surface, to belong to their culture. It is this surface-level plausibility that makes the deeper issue easy to overlook. When a culturally adapted word problem looks correct, there is little reason to ask what kinds of cultural judgments produced it, what alternatives were excluded, or whether another model would have imagined the same local world.

Our study begins from that concern. We do not treat LLMs as substitutes for the situated expertise that culturally responsive pedagogy requires. Instead, we argue that because this is skilled pedagogical work, any system increasingly used to assist with it deserves careful scrutiny. If educators rely on LLMs to help draft adapted materials, then the outputs must be evaluated not only for grammaticality and fluency but also for the cultural assumptions they carry. The question is not whether educators should remain central to adaptation; they should. The question is whether the LLMs that are being used are stable, diverse, and pedagogically trustworthy enough across the languages and regions on a systematic level to be useful in educational practice.

This paper explores the following research questions. The first concerns consistency. \textbf{RQ1. When the same mathematics word problem is given to multiple frontier LLMs with the same instruction (e.g., translate and culturally adapt this for Sicilian speakers), do they produce the same cultural output?} If they do not, then model choice is not merely a technical decision about system performance or language quality. It is also a cultural decision, because the selected model shapes the world of names, places, objects, practices, and institutions that students are asked to inhabit. The second question concerns cultural variety across the adaptation process itself. \textbf{RQ2. When English math word problems are adapted into another language, do the outputs preserve the range of cultural entities present in the source set, or does adaptation paradoxically compress that variety into a narrower and more homogeneous target world?} A system that appears to localize may, at scale, repeatedly select the same kinds of names, foods, places, and practices, thereby reducing rather than expanding cultural representation. The third question concerns how models allocate cultural attention. Cultural adaptation is not a single act but a sequence of choices about what to preserve, what to localize, what to generalize, and what to ignore. \textbf{RQ3. Which entities do LLMs appear to treat as carrying the most cultural weight?} And do those model-internal priorities resemble the judgments a teacher would make when adapting a problem for real students in a specific classroom?

To answer these questions, we audit how three frontier LLMs, Claude Opus 4, GPT-4.1, and Gemini 2.5 Pro, culturally adapt 60 English GSM8K math word problems into seven target languages across three regions: Bengali, Hindi, and Punjabi for India; Urdu and Sindhi for Pakistan; and Italian and Sicilian for Italy. For each of the 1,260 resulting translations, we identify culturally adaptable entities in the English source, including names, foods, currencies, places, institutions, and practices, and align each entity with its realization in the target-language output. This produces an entity-level dataset of 6,489 transformations spanning 66 fine-grained entity types organized into 26 super-categories (Appendix I), annotated with one of five action labels: \textit{preserved}, \textit{localized}, \textit{generalized}, \textit{type changed}, or \textit{missing}. Using this dataset, we measure cross-model agreement on both transformation type and specific substitution (RQ1), quantify entropy collapse between the diversity of English source entities and their translated counterparts (RQ2), and characterize which entity categories models treat as culturally salient versus functionally universal, as well as how these choices vary across languages and regions (RQ3). Beyond aggregate patterns, we surface case studies, such as Bengali translations defaulting to Bangladeshi taka for students in India and Easter egg hunts being reframed as Eid activities, to show how surface-level fluency can mask deeper cultural misattribution.

\section{Related Work}
Cultural context has a powerful effect on how students engage with math word problems for two reasons. The first is cognitive. Students must build a mental model of the situation a problem describes \citep{kintsch1985understanding, cummins1988role}, and effort spent interpreting an unfamiliar setting leaves less working memory for the reasoning the problem is designed to teach \citep{sweller1988cognitive, chandler1991cognitive}. The second is about belonging. When a problem assumes a world a student does not recognize, the mismatch is not only a comprehension cost but a signal that the materials were written for someone else, and a sense of not belonging in academic contexts is well-documented to depress motivation and achievement \citep{walton2007questionbelonging, good2012why}. For this reason, localization is not merely translation. Translation changes language; localization changes culturally meaningful referents, such as names, currencies, foods, settings, and practices, while preserving mathematical structure. Done well, it draws on community knowledge and lived experience as resources for learning \citep{ladson1995toward, gay2018culturally}. It is also pedagogical work that resists shortcuts. \citet{hammond2014culturally} distinguishes surface culture, the observable layer of food, dress, and names, from deep culture, the unspoken norms, registers, and worldviews that shape how people interact. Substituting a local name into an otherwise unchanged problem stays on the surface; understanding which traditions, practices, and registers belong in a child's everyday world reaches the deeper layer. This need for contextual knowledge is consistent with funds-of-knowledge research, which emphasizes that meaningful instruction draws on students' household, community, and cultural resources \citep{moll2006funds, gonzalez2001bridging}. This is part of what makes manual cultural adaptation so demanding and difficult to reproduce at scale.

LLMs are increasingly being used to scale this task at the production level. A 2024 OECD survey reports that 37\% of teachers across member countries already use generative AI for work-related tasks \citep{oecd2026outlook}, and adaptation is happening institutionally as well: the National Council of Educational Research and Training in India has used AI/ML to translate Grade 1-2 textbooks from English into 22 Indian languages \citep{pib2026aieducation}, and Khanmigo, Khan Academy's GPT-4-based teacher and student assistant, launched free for all Indian teachers in Hindi in 2024 \citep{khanacademy2024india}. UNESCO's \emph{Guidance for Generative AI in Education} warned that such deployments threaten ``linguistic and cultural diversities'' and noted that fewer than 10\% of educational institutions had formal AI guidance at the time of publication \citep{miao2023unesco}.

LLMs do not adapt culture in neutral ways. \citet{naous2024having} show that models generating text for Arab cultural settings can insert culturally inappropriate entities, such as references to alcohol, and \citet{wang2024not} find that even non-English prompts can elicit English-culture defaults (like Thanksgiving references), both examples of a deeper tendency where fluent, superficially plausible text still carries assumptions from a different culture. \citet{li2024culture} show that LLMs treat Western culture as the unmarked default and mark non-Western cultures with linguistic cues like `traditional.' Across multilingual evaluations, LLMs consistently align more closely with Western cultural norms than with target-language communities, and AI suggestions can push users toward more Westernized expression, flattening rather than reflecting non-Western cultures \citep{tao2024cultural, cao2023assessing, arora2023probing, durmus2023towards, agarwal2025ai, ryan2024unintended}. The pedagogical work of cultural adaptation, deep, situated, grounded in lived experience, is being delegated to systems that default to a different culture entirely.

Recent work has begun to ask whether math word problems are culturally neutral for LLMs. \citet{karim2025lost} ask a focused question: if the same GSM8K problem is rewritten into different cultural settings while keeping the underlying mathematics unchanged, do models still handle reasoning in the same way? GSM8K is a dataset of 8.5K high-quality grade school math word problems \citep{cobbe2021training}. They create culturally rewritten versions by changing contextual elements such as names, foods, and places while preserving the numerical logic and language, and find that cultural context can affect how well LLMs solve math word problems. \citet{tomar2025mathematics} ask a related but broader question: is GSM8K itself already culturally situated rather than culturally neutral? They create region-specific variants for Africa, India, China, Korea, and Japan, using entity and scenario perturbations while keeping the mathematical content aligned, and find that models generally perform best on the original U.S.-centric versions. \citet{abebe2025bridging} move closest to our setting by treating socio-cultural localization of math word problems in low-resource languages as a distinct task rather than a simple translation problem; their LLM-driven localization pipeline replaces English-centric entities with local alternatives and reveals performance gaps that translation-only benchmarks hide.

Our work differs from \citet{karim2025lost} and \citet{tomar2025mathematics} by shifting attention away from downstream answer accuracy and toward the cultural adaptation process itself. \citet{abebe2025bridging} is similar to ours in that it also treats cultural adaptation as a problem in its own right and recognizes that meaningful localization requires more than literal translation, but our work differs in both scope and goal. They focus on whether localization can be automated as a pipeline for low-resource benchmarking. By contrast, we focus on the adaptation process itself: which cultural entities models change, how those values shift across models, languages, and regions, and whether the resulting adaptations reflect deeper cultural grounding or mostly surface-level substitution. In that sense, their work asks whether LLMs can localize math word problems, while ours asks what kind of culture LLMs actually produce when they do.

\section{Data Generation Methods and Datasets}
Our analysis proceeds in three stages. We first select a set of culturally rich English source problems. We then generate culturally adapted translations of these problems across languages and models. Finally, we convert the resulting translation corpus into an entity-level dataset that records how each source entity is transformed in the target-language output.

\textbf{Source Problem Selection.} \label{sec:source_problem_selection}
We begin with the dataset introduced by \citet{karim2025lost}, which is derived from GSM8K \citep{cobbe2021training} and annotates English grade-school math word problems for culturally adaptable entities such as foods, sports, and names. These annotations span 52 cultural entity categories. From this resource, we sample 60 culturally rich English source problems that together cover all 52 categories.

\textbf{Multilingual Cultural Translation Corpus.} \label{sec:translation_corpus}
We translate each of the 60 source problems into seven target languages across three regional contexts: Hindi, Bengali, and Punjabi for India; Urdu and Sindhi for Pakistan; and Italian and Sicilian for Italy. For each target language, translations are generated by three frontier language models: Claude Opus 4, GPT-4.1, and Gemini 2.5 Pro.

The prompt is framed from a teacher's perspective:
\begin{quote}
\small
You are an elementary school math teacher in [country], teaching students in [language]. Translate the following math word problem from English into [language] and adapt the problem so that it fits the cultural context of students in [country].
\end{quote}

No explicit instructions are given about which entities to modify or preserve. In total, translating 60 source problems across 3 models and 7 target languages yields 1{,}260 translated problems. We set temperature to 0.7 for all generations to encourage lexical variation while maintaining coherence \citep{singh2023hide, moreno2024exploring}, and keep all other parameters at their default values.

\textbf{Entity-Level Dataset.}
We next convert the translation corpus into an entity-level dataset. For each English source problem, we identify entities that could plausibly be preserved, localized, generalized, changed, or omitted during translation. We use the term \textit{entity} broadly to refer to a lexical item or short span denoting an identifiable element in the problem narrative, including people, places, institutions, events, materials, roles, and other concrete or abstract noun-like concepts. This definition is broader than explicitly cultural content: rather than assuming in advance which entities models will treat as culturally salient, we annotate a wide range of source entities and observe how they transform.

Although \citet{karim2025lost} provides the starting point for source problem selection, we conduct a new round of source-side entity annotation because our schema is broader than the cultural categories in that work. The final source-side schema contains 66 fine-grained entity types and 26 super entity types. A running example is shown at the end of this section.

\textbf{Target-Side Entity Alignment.}
For each translated problem, we align every source-side entity with its realization in the target-language output. Because the corpus contains 6,489 source--target entity alignments across seven languages, fully manual span search would be infeasible. We use an LLM-assisted mapping procedure with GPT-4.1 to identify candidate target-side surface forms. The model is given the English source problem, the source entity inventory, and the translated problem, and is asked to return each source entity paired only with the value that explicitly appears in the translation. This step is used only to narrow the search space: GPT-4.1 proposes candidate spans but does not assign action labels.

Each candidate alignment is then manually validated. We first check that the mapped output preserves the authoritative source entity keys: every source key must appear exactly once, and candidate outputs cannot add, remove, merge, or rename keys. We also check that the number of mapped target values matches the number of source values for each key. Outputs failing these checks are inspected. Incorrect alignments, missing entities, and over or under specified spans are also manually corrected. For ambiguous cases, we consulted lab members familiar with the relevant language or regional context.

We also record a standardized English backtranslation of each translated entity value using Google Translate. This provides a shared cross-lingual representation for comparing substitutions across languages and models.

\textbf{Action Label Assignment.} Given the source-side entities and their values, the corresponding target-side entities and values, and our goal of recording how the LLM transformed each source entity in translation, we assign one of five action labels to each reviewed row: \textbf{preserved} (value and entity type identical to source); \textbf{localized} (entity type unchanged, value replaced with a locally familiar equivalent, we do not judge appropriateness); \textbf{generalized} (entity type unchanged, but the translated value is semantically broader or less specific; takes precedence over \textit{localized} when both apply); \textbf{type changed} (both value and entity type differ from source; takes precedence over \textit{generalized} when both apply); \textbf{missing} (no explicit realization of the source entity appears in the translation, including cases where at least one value in a multi-value entity type goes unrealized).

This process produces an entity-level dataset of 6,489 transformations, which we use to compare how specific entities change across languages, regions, and models.

\subsection{Running Example of Entity-Level Annotation}
\label{app:example_source_side_entities}
As a running example, consider the following source problem:
\begin{quote}
\small
The tooth fairy left Sharon \$5.00 in exchange for the first tooth Sharon lost. Then, the tooth fairy gave Sharon \$1.00 for each of the next three teeth Sharon lost. And for each of the last 2 teeth Sharon lost, the tooth fairy gave Sharon half the amount of money per tooth as Sharon had received for each of the previous three teeth. How much money did the tooth fairy leave Sharon, in dollars?
\end{quote}
For one Sicilian translation generated by Claude, the problem was adapted as follows:
\begin{quote}
\small
A Fatina di li denti lassau a Carmela 5 euro pi lu primu denti ca pird\`{i}u. Poi, a Fatina ci detti 1 euro pi ognunu di l'autri tri denti ca pird\`{i}u. E pi l'urtimi du denti ca pird\`{i}u, a Fatina ci detti a met\`{a} di li sordi pi ogni denti rispettu a chiddu ca avia ricivutu pi ognunu di li tri denti prima. Quanti sordi ci lassau a Fatina a Carmela, in euro?
\end{quote}
Table~\ref{tab:running_example_pipeline} shows how source-side entities are aligned to target-side realizations and assigned action labels.
\begin{table*}[t]
\centering
\small
\begin{tabular}{llllll}
\hline
\textbf{Source type} & \textbf{English value} & \textbf{Target type} & \textbf{Target value} & \textbf{Backtranslation} & \textbf{Action} \\
\hline
person & Sharon & person & Carmela & Carmela & localized \\
currency\_name & dollars & currency\_name & euro & euro & localized \\
currency\_symbol & \$ & currency\_name & euro & euro & type changed \\
mythical\_character & tooth fairy & mythical\_character & Fatina di li denti & tooth fairy & preserved \\
\hline
\end{tabular}
\caption{Running example of the entity-level annotation pipeline for one Sicilian Claude translation. The table shows source entities, target-language realizations, normalized English backtranslations, and assigned action labels. The currency symbol row shows \textit{type changed}: the source entity is a symbol, but the translated realization is a currency name.}
\label{tab:running_example_pipeline}
\end{table*}
\subsection{Annotator Background and Inter-Annotator Reliability}
The full entity-level dataset was coded by the primary annotator, who has native proficiency in Urdu and Sindhi, spoken and listening proficiency in Hindi and Punjabi, and can read romanized Hindi and Punjabi. The primary annotator does not have proficiency in Bengali, Italian, or Sicilian.

To assess reliability, we drew a 15\% stratified sample by language and model, yielding 966 entity rows. An independent second annotator coded this sample using the same codebook. The second annotator has native proficiency in Hindi and spoken proficiency in Punjabi and Urdu, and can read romanized Punjabi and Urdu. The second annotator does not have proficiency in Bengali, Italian, Sicilian, or Sindhi.

Raw agreement was 0.910 and Cohen's $\kappa$ was 0.859, with a bootstrap 95\% confidence interval of [0.831, 0.886]. Language-level reliability ranged from $\kappa=0.735$ for Sicilian to $\kappa=0.909$ for Bengali, and model-level reliability ranged from $\kappa=0.824$ for Gemini to $\kappa=0.881$ for Claude.

Because annotator language expertise varied across target languages, our annotation task was designed to record observable entity transformations rather than independent judgments of linguistic fluency or cultural appropriateness. For ambiguous cases, we consulted lab members familiar with the relevant language or regional context, including cases involving Bengali, Italian, and Sicilian.

\section{Analysis Framework}
To address our research questions, we organize the analysis around four metrics.
\textbf{Cross-Model Agreement (RQ1).} For each of the 2,163 entity-language instances (309 entities mentions $\times$ 7 languages), we measure cross-model agreement at two levels. First, we compute three-way agreement separately for action and value: the proportion of instances for which all three models assign the same action label, and the proportion for which all three models produce the same normalized back-translated substitution. Second, we compute pairwise agreement for each model pair (Claude–Gemini, Claude–GPT, Gemini–GPT) as the proportion of instances on which the two models agree.
\textbf{Value Entropy Collapse (RQ2).} For each super entity type, we compute Shannon entropy~\citep{shannon1948mathematical} over the English source values and compare it to the entropy of corresponding back-translated values for each language and model combination. Source and target entropy are computed over the same matched observations, excluding entities labeled \textit{missing} from both sides. We additionally use the Miller--Madow \citep{paninski2003estimation} correction to account for small-sample bias and paired bootstrap resampling (2,000 draws) to obtain 95\% confidence intervals. A negative entropy difference indicates a reduction in value diversity after translation, while a positive difference indicates increased diversity. We aggregate across super entity types to identify systematic loss in diversity by language and model.
\textbf{Adaptation Balance (RQ3).} We compute an adaptation balance score for each entity type: the probability of localization minus the probability of preservation, ranging from -1 (always preserved) to +1 (always localized). Entity types with scores near zero occupy a contested middle where neither action dominates. We report these scores by entity type and aggregate to the super entity type level to characterize which entities models treat as culturally marked versus functionally universal.
\textbf{Cross-Linguistic Similarity (RQ3).} We compute Jensen-Shannon \citep{lin2002divergence, endres2003new} divergence between each language pair based on how they distribute actions across entity types. Languages with low divergence handle entities similarly; high divergence indicates systematically different strategies.

\section{Results and Discussion}
We organize our findings around the three research questions. Across all analyses, a single tension recurs: models produce outputs that appear locally appropriate while concealing instability, cultural compression, and shallow substitution. We discuss case studies of cross-cultural and cross-regional contamination that illustrate how this surface plausibility can mask deeper failures.

\subsection{RQ1. Cross-Model Consistency}
\label{sec:results_RQ1}
\begin{figure}[t]
\centering
\includegraphics[width=\columnwidth]{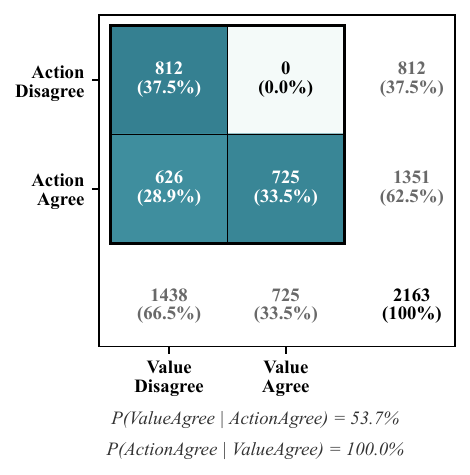}
\caption{Joint distribution of three-way agreement on action (transformation type) and value (specific substitution) across all 2,163 entity–language instances. Each cell shows the number and percentage of instances falling into that combination of action and value agreement. All three models agree on the action in 62.5\% of instances and on both action and value in 33.5\%. The top-right cell is empty by construction: if all three models produce the same back-translated value, they have necessarily applied the same action. Conditional probabilities below the matrix summarize the asymmetric relationship between the two forms of agreement.}
\label{fig:action_value_agreement}
\end{figure}

Models do not produce consistent cultural outputs. Across 2,163 entity-language combinations (309 entities mentions $\times$ 7 languages), all three models agree on the action in 62.5\% of cases but agree on both action and output value in only 33.5\% of cases. Models often converge on the same transformation type while choosing different specific substitutes. When adapting an English name like ``Sarah'' for Italian students, all three models may agree on localization yet choose different names: ``Giulia,'' ``Sofia,'' or ``Chiara.'' Among cases where models agree on an action, they agree on output value only 53.7\% of the time (Figure~\ref{fig:action_value_agreement}).

\begin{figure}[t]
\centering
\includegraphics[width=\columnwidth]{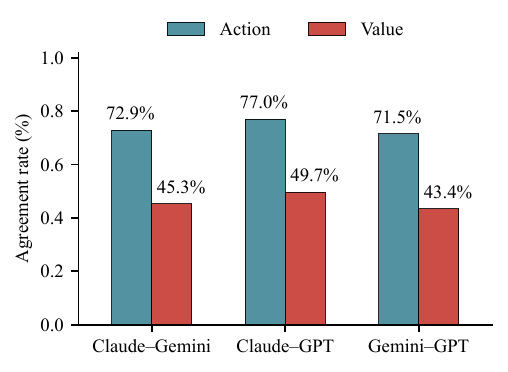}
\caption{Pairwise agreement rates on action, defined as transformation type, and value, defined as the specific substitution. Claude and GPT are most similar (77.0\% action agreement, 49.7\% value agreement), followed by Claude and Gemini (72.9\%, 45.3\%), with Gemini and GPT least similar (71.5\%, 43.4\%)}
\label{fig:model_agreement}
\end{figure}

\begin{figure*}[t]
\centering
\includegraphics[width=\textwidth]{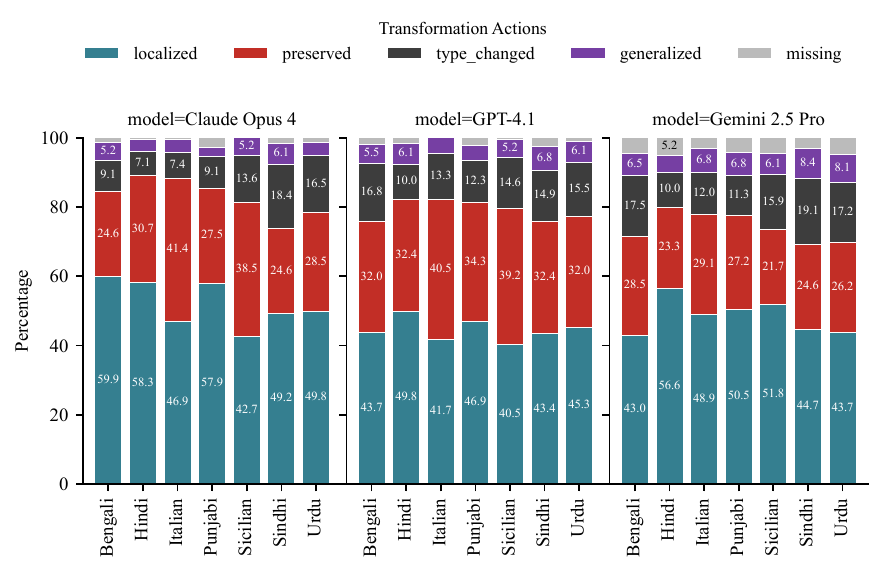}
\caption{Distribution of the five transformation actions (localized, preserved, type-changed, generalized, missing) across the seven target languages, separately for each model. Percentages are computed over all entity–language instances. Claude localizes most aggressively across languages, GPT-4.1 preserves most conservatively, and Gemini 2.5 Pro shows the highest type-changed rates.}
\label{fig:action_distribution}
\end{figure*}

Pairwise comparisons (Figure~\ref{fig:model_agreement}) reveal systematic differences. Claude and GPT are most similar (77\% action agreement, 49.7\% value agreement), followed by Claude and Gemini (72.9\%, 45.3\%), with Gemini and GPT least similar (71.5\%, 43.4\%). Figure~\ref{fig:action_distribution} shows why: Claude localizes most aggressively across languages, GPT preserves most conservatively, and Gemini shows the highest type-changed rates, restructuring entity categories more frequently. These patterns recur across all seven target languages, suggesting stable model-specific tendencies rather than language-specific variation.

\begin{figure}[t]
\centering
\includegraphics[width=\columnwidth]{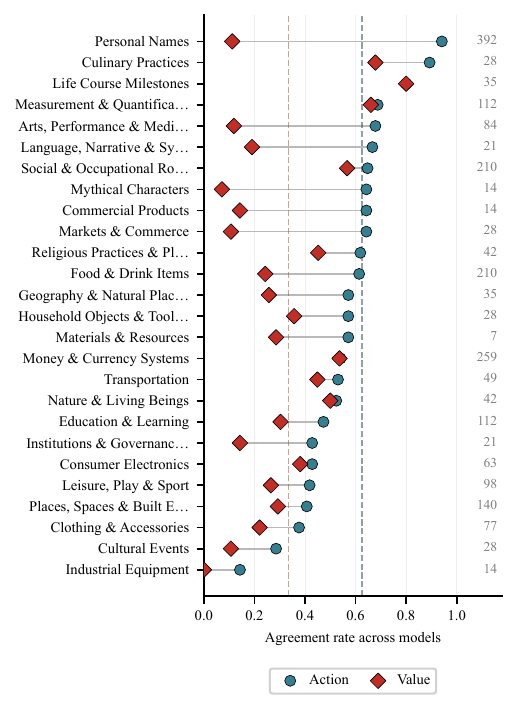}
\caption{Three-model agreement by entity super-category. Points show action and value agreement rates; connecting lines show the gap between agreement on transformation type and agreement on the specific substituted value. Personal Names show high action agreement but lower value agreement, while Industrial Equipment, Cultural Events, and Clothing \& Accessories show low agreement even on actions.}
\label{fig:super_entity_dumbbell}
\end{figure}

Agreement also varies by entity category (Figure~\ref{fig:super_entity_dumbbell}). Personal Names produce the strongest convergence on action but weaker convergence on specific values, while Industrial Equipment, Cultural Events, and Clothing \& Accessories show low agreement even on actions. Across nearly all categories, action agreement exceeds value agreement.

Taken together, these findings show that model choice is a cultural decision, not merely a technical one. An educator switching from Claude to Gemini, or from GPT to Claude, is not simply changing system performance or language quality. They are also changing the names, places, foods, institutions, and practices that students encounter in adapted math problems. Even when models agree on the broad form of adaptation, they often instantiate it through different values.

Disagreement among models is not in itself a failure. For many entity types, multiple substitutions can be culturally valid and lexical variation across models could equally be read as healthy diversity in a space without a single correct answer. We treat low value agreement as significant not because disagreement is inherently bad, but because it combines with the entropy collapse documented in RQ2: models disagree with one another while each individually drawing from a narrow pool. The concern is the joint pattern, not either finding alone.

\subsection{RQ2. Does Adaptation Compress Cultural Variety?}
\label{sec:results_RQ2}

\begin{figure}[t]
\centering
\includegraphics[width=\columnwidth]{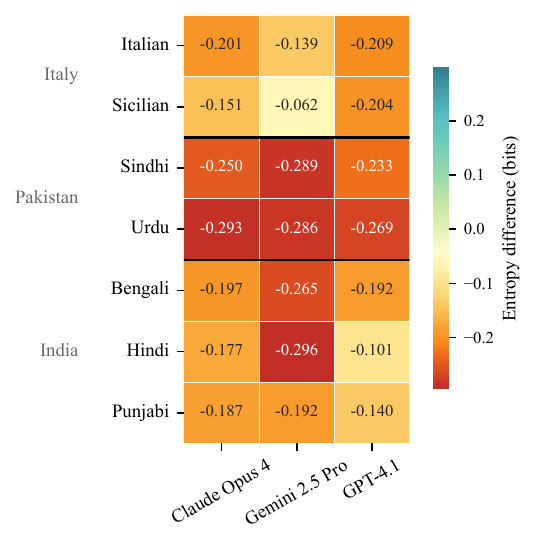}
\caption{Mean entropy difference between English source entity values and model-generated back-translated values, disaggregated by language and model. Negative values indicate entropy collapse, where outputs draw from a narrower set of entity values than the English source. Reductions range from 0.062 bits (Sicilian with Gemini) to 0.296 bits (Hindi with Gemini, with Urdu under Claude nearly identical at 0.29).}
\label{fig:entropy_by_lang_model}
\end{figure}

Cultural adaptation systematically reduces the diversity of cultural referents. All 21 language and model combinations show negative entropy differences (Figure~\ref{fig:entropy_by_lang_model}). Reductions range from 0.062 bits (Sicilian with Gemini) to 0.296 bits (Hindi with Gemini). On average the three models are close (Gemini $-0.22$ bits, Claude $-0.21$, GPT $-0.19$), but they differ in how steady they are across languages: Gemini produces both the deepest and the shallowest reductions in the corpus, while Claude stays within a narrower band. Pakistani languages, Urdu and Sindhi, show the deepest collapse, Indian languages show moderate collapse, and Italian and Sicilian show the smallest reductions.

\begin{figure}[t]
\centering
\includegraphics[width=\columnwidth]{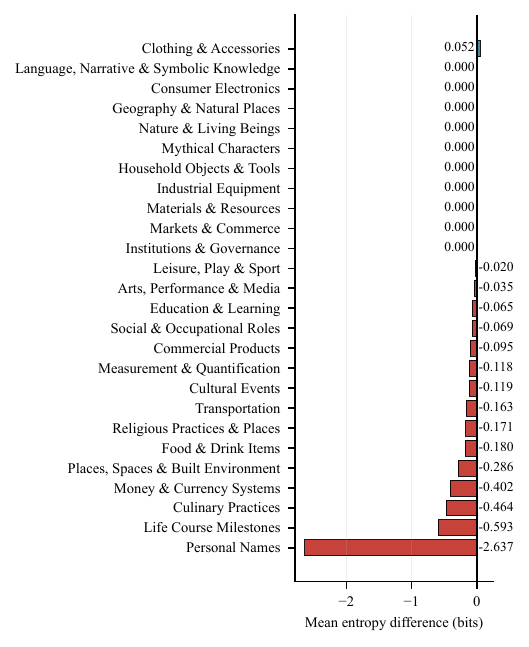}
\caption{Entropy collapse by super-category. Bars show mean entropy difference between English source values and back-translated model outputs; negative values indicate collapse, while positive values indicate maintenance or expansion. Personal Names show the largest reduction (2.637 bits), followed by Life Course Milestones (0.593) and Culinary Practices (0.464).}
\label{fig:value_entropy}
\end{figure}

Collapse is concentrated in specific entity types. Personal Names show the strongest reduction at 2.64 bits, far larger than any other category (Figure~\ref{fig:value_entropy}). Life Course Milestones (0.59 bits) and Culinary Practices (0.46 bits) show moderate reductions. Many categories remain near zero. Only Clothing \& Accessories shows a positive mean (0.052 bits), driven by a small number of cells in which one additional distinct value appears.

Entropy collapse comes from two sources. A few categories collapse everywhere, regardless of language or model, while others collapse only in certain regions. Personal Names is the clearest case of the first kind: it shows strong negative values in almost every language and model, which is why it dominates the overall pattern. Life Course Milestones and Culinary Practices collapse just as consistently at a smaller scale. Money \& Currency Systems is the clearest case of the second kind. It collapses sharply in Pakistani settings, where Sindhi and Urdu fall between $-0.87$ and $-0.94$ across all three models, but stays near zero in Hindi and Punjabi and turns slightly positive in Italian and Sicilian under Claude and Gemini. The full category, language, and model heatmap is in Appendix II.

These results remain consistent after accounting for small sample sizes and uncertainty. Because standard entropy estimates can be biased when there are few observations, we repeated the analysis using the Miller--Madow correction. The overall mean entropy difference became slightly more negative, from $-0.206$ to $-0.250$ bits, and remained negative for all 21 language--model combinations. This suggests that the overall pattern is not caused by small-sample bias. Bootstrap confidence intervals also support the result: 19 of the 21 language--model combinations have intervals entirely below zero. The two exceptions, Italian and Sicilian under Gemini 2.5 Pro, are also the combinations with the smallest observed reductions. Small categories are less informative: 82\% of cells with $n \leq 5$ show no entropy difference, compared with only 6\% of cells with $n > 20$. Thus, a zero difference in a small category does not provide strong evidence that diversity was preserved. Per-combination estimates and intervals are reported in Appendix II.

These results show a tension in how models perform cultural adaptation. Models often change entity values to make problems seem locally appropriate, but across the full corpus they repeatedly draw from a limited set of substitutes. As a result, the translated outputs may look culturally appropriate in individual examples while becoming more homogeneous overall. The same names, foods, and places appear again and again, reducing cultural diversity at the corpus level.

\begin{figure}[t]
\centering
\includegraphics[width=\columnwidth]{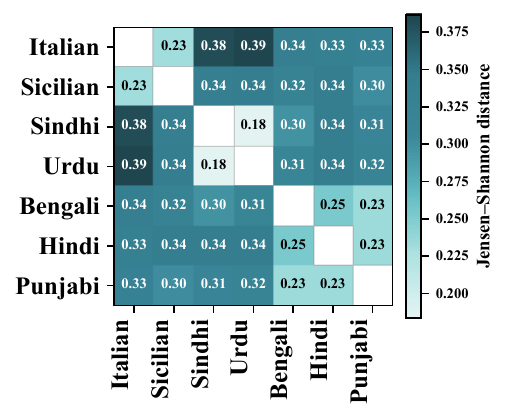}
\caption{Jensen-Shannon distance between language pairs based on entity-type action distributions. Lower values indicate more similar adaptation strategies. Italian and Sicilian show nearly identical patterns (JSD = 0.23), Sindhi and Urdu are the most similar pair in the dataset (JSD = 0.18), and Bengali, Hindi, and Punjabi form a tight cluster (JSD = 0.23–0.25)}
\label{fig:jsd_heatmap}
\end{figure}

\subsection{RQ3.Which Entities Do LLMs Treat as Culturally Salient?}

\label{sec:results_RQ3}
Models do not treat all entity types as equally deserving of cultural adaptation. Table~\ref{tab:adaptation_spectrum_localized} shows entity types where localization dominates. Performance contexts, newspapers, and music items lead the list with localization rates at or above 0.95, followed by currency names and denominations. Personal names show similarly high localization (balance = 0.916) and represent by far the largest sample. Food items, markets, and religious places also show strong localization tendencies.

\begin{table}[t]
  \small
  \begin{tabular}{lrr}
    \hline
    \textbf{entity\_type} & \textbf{N} & \textbf{balance} \\
    \hline
    performance\_context & 21 & 1.000 \\
    newspaper & 42 & 1.000 \\
    music\_item & 42 & 0.952 \\
    currency\_denomination & 21 & 0.952 \\
    currency\_name & 105 & 0.943 \\
    person & 1176 & 0.916 \\
    market\_type & 42 & 0.858 \\
    religious\_place & 21 & 0.810 \\
    food\_item & 126 & 0.786 \\
    residence & 21 & 0.715 \\
    sport\_or\_game\_event & 21 & 0.715 \\
    movie & 21 & 0.667 \\
    language & 63 & 0.650 \\
    online\_marketplace & 42 & 0.643 \\
    governing\_bodies & 21 & 0.619 \\
    dance\_style & 84 & 0.560 \\
    \hline
  \end{tabular}
  \caption{\label{tab:adaptation_spectrum_localized}
  Cultural adaptation spectrum: localized end (\textit{adaptation balance} \(> 0.5\)), ranked by adaptation balance. N is the number of observations (7 languages \(\times\) 3 models per source occurrence); rows with N=21 represent a single source problem and should be interpreted with caution. See Appendix II for preserved and localized rates per row.}
\end{table}

\begin{table}[t]
  \small
  \begin{tabular}{lrr}
    \hline
    \textbf{entity\_type} & \textbf{N} & \textbf{balance} \\
    \hline
    vehicle\_part & 21 & -1.000 \\
    meal\_type & 21 & -1.000 \\
    plant\_or\_flower & 21 & -1.000 \\
    participant\_role & 105 & -0.980 \\
    life\_event & 105 & -0.895 \\
    season & 21 & -0.667 \\
    question\_type & 63 & -0.556 \\
    electronics & 189 & -0.556 \\
    \hline
  \end{tabular}
  \caption{\label{tab:adaptation_spectrum_preserved}
  Cultural adaptation spectrum: preserved end (\textit{adaptation balance} \(< -0.5\)), ranked by adaptation balance. N is the number of observations (7 languages \(\times\) 3 models per source occurrence); rows with N=21 represent a single source problem and should be interpreted with caution. See Appendix II for preserved and localized rates per row.}
\end{table}

\begin{table}[t]
  \small
  \begin{tabular}{lrr}
    \hline
    \textbf{entity\_type} & \textbf{N} & \textbf{balance} \\
    \hline
    product\_category & 42 & -0.286 \\
    book\_type & 147 & -0.286 \\
    occupation\_role & 210 & -0.229 \\
    appliance & 21 & -0.142 \\
    activity & 84 & -0.131 \\
    retail\_venue & 210 & -0.129 \\
    organization & 42 & -0.096 \\
    class\_type & 42 & -0.071 \\
    measurement\_unit & 336 & -0.062 \\
    dessert & 21 & -0.048 \\
    sport\_or\_game & 84 & -0.047 \\
    natural\_feature & 63 & -0.032 \\
    mythical\_character & 42 & 0.000 \\
    cooking\_method & 63 & 0.000 \\
    workplace & 63 & 0.016 \\
    animal & 63 & 0.079 \\
    performance\_type & 21 & 0.095 \\
    animal\_feed & 21 & 0.095 \\
    beverage & 105 & 0.114 \\
    toy\_or\_game\_item & 105 & 0.115 \\
    event & 84 & 0.131 \\
    baked\_good & 189 & 0.132 \\
    construction\_material & 21 & 0.142 \\
    entertainment\_venue & 42 & 0.143 \\
    learning\_institution & 42 & 0.167 \\
    jewelry & 21 & 0.191 \\
    country & 21 & 0.238 \\
    scent\_type & 42 & 0.238 \\
    apparel & 168 & 0.250 \\
    restaurant & 21 & 0.286 \\
    currency\_symbol & 651 & 0.287 \\
    \hline
  \end{tabular}
  \caption{\label{tab:adaptation_spectrum_contested}
  Cultural adaptation spectrum: contested middle (\(\left|\textit{adaptation balance}\right| < 0.3\)), ranked by adaptation balance. N is the number of observations (7 languages \(\times\) 3 models per source occurrence); rows with N=21 represent a single source problem and should be interpreted with caution. See Appendix for preserved and localized rates per row.}
\end{table}

Table~\ref{tab:adaptation_spectrum_preserved} shows the opposite pole: vehicle parts, meal types (e.g., breakfast), and plants or flowers are preserved in 100\% of cases, with participant roles (e.g., student) and life-course milestones (e.g., university admission, wedding) close behind, with preservation rates of 0.99 and 0.93, respectively.

Between these poles lies a contested middle (Table~\ref{tab:adaptation_spectrum_contested}), where models do not consistently prioritize entities for cultural adaptation. Occupations, retail venues, measurement units, and currency symbols sit near balance = 0, with no clear default toward either preservation or localization. Full versions of all three tables are provided in Appendix II.

These dimensions match documented approaches to culturally relevant word problems, where teachers adapt names, settings, activities, foods, family references, and celebrations \citep{herron2009culturally}, while organizations are adapted less often than surface cultural features such as customs \citep{fan2018cultural}. Models show both patterns. This also fits evidence on cognitive load: \citet{ntumi2026culturally} found that culturally unfamiliar word-problem contexts increased cognitive load, while familiar contexts improved clarity and engagement.

The adaptation patterns reveal a gap between surface-level localization and deeper cultural responsiveness. Hammond's Culture Tree framework distinguishes surface culture (food, dress) from deep culture (values, worldview) \citep{hammond2014culturally}, and the National Council of Teachers of Mathematics cautions that instruction must move beyond surface markers like student names, cultural food, and festivities \citep{nctm2024culture, desai2022ethnomodeling}. Models focus almost entirely on the surface layer. They change names, foods, and currencies while preserving grade-level designations that embed culturally specific structures, terms like ``third-graders'' are preserved universally, even though Italian students attend ``terza elementare,'' Pakistani students ``Class 3,'' and Indian students ``Standard 3.''

This surface-level substitution can produce culturally incoherent content when models lack cross-cultural understanding of which activities belong to which cultural practices. A case from our dataset illustrates this: models replaced an Easter egg hunt with an ``egg search competition on Eid'' for Urdu and Sindhi translations, substituting the holiday name while preserving the activity structure. Egg hunts are not an Eid tradition. The substitution demonstrates that models sometimes apply pattern-matching rules (replace Western holiday with regional holiday) without cultural knowledge of what practices accompany those holidays, creating adaptations that appear localized on the surface but remain culturally inauthentic.

While the adaptation spectrum reveals which categories models prioritize universally, examining cross-linguistic divergence shows that models apply region-specific cultural knowledge rather than treating each target language independently. Figure~\ref{fig:jsd_heatmap} shows that adaptation patterns cluster by geographic region: Italian and Sicilian show nearly identical patterns (JSD = 0.23), Sindhi and Urdu are the most similar pair in the dataset (JSD = 0.18), and Bengali, Hindi, and Punjabi form a tight cluster (JSD = 0.23–0.25). Cross-regional comparisons show substantially higher divergence, with European and South Asian languages differing by 0.30 to 0.39. This geographic clustering indicates that models have internalized region-specific cultural schemas rather than applying universal adaptation rules to each language independently. Appendix II decomposes this divergence by entity type, showing that personal names are handled consistently across languages while categories such as animal feed and governing bodies are treated inconsistently.

However, regional knowledge is not always accurate. Bengali provides a revealing example of cross-regional bias. Although the prompt explicitly specified ``an elementary school math teacher in India, teaching students in Bengali,'' models used the Bangladeshi taka in 76.2\% of Bengali currency instances (77 of 101), whereas Hindi translations used Indian rupees in 100\% of comparable cases. The same pattern appeared in broader cultural substitutions: when replacing Western songs, one Bengali adaptation produced ``Amar Sonar Bangla,'' the national anthem of Bangladesh. This example does not explain all cross-language differences in the dataset, but it shows one clear failure mode. Even when the prompt specifies Bengali for students in India, models sometimes default to associations with Bangladesh instead. For teachers, this means that a problem can look culturally adapted on the surface while still placing students in the wrong regional setting.

\section{Limitations}
This study has several limitations. First, our analysis operates at the entity level, which allows us to systematically compare how models transform names, foods, currencies, places, and other cultural referents across languages. However, math word problems are not simply collections of independent entities. They are short narrative worlds in which entities interact with one another and the surrounding context. An entity that appears appropriate in isolation may still become culturally incoherent or pedagogically awkward when interpreted within the full problem. Our results, therefore, capture how models transform individual cultural referents, but not the broader coherence or educational impact of those transformations in context.

Second, our annotation scheme is descriptive rather than normative. We record whether an entity was preserved, localized, generalized, type-changed, or missing, but we do not evaluate whether a substitution is culturally appropriate, authentic, or pedagogically desirable for the target setting. This was a deliberate design choice. The dataset spans seven languages, three regional contexts, and many culturally specific domains, making consistent judgments of cultural correctness difficult to validate at scale. Annotator language expertise also varied across the target languages: the primary annotator has native proficiency in Urdu and Sindhi and partial proficiency in Hindi and Punjabi, while the second annotator has native proficiency in Hindi and partial proficiency in Punjabi and Urdu. Neither annotator has proficiency in Bengali, Italian, or Sicilian. We therefore use backtranslation, explicit target-side surface matching, and consultation with lab members familiar with the relevant language or regional context to support cross-lingual entity comparison. The action labels should not be read as judgments of cultural appropriateness or translation quality.

Third, our findings are based on a single prompting setup and fixed generation parameters. We use one instruction template designed to simulate a teacher adapting a problem for students in a given language and country, but we do not test how outputs change under alternative prompts, temperatures, system messages, or multi-step adaptation workflows. The results should therefore be interpreted as patterns under one controlled setup rather than as exhaustive descriptions of each model's behavior.

Finally, the model snapshots were captured in mid-2025. In a rapidly changing field, newer versions of Claude, GPT, and Gemini may produce different adaptation patterns. Whether newer models reduce or reproduce the instabilities, regional misattributions, and diversity collapse documented here remains an empirical question.

\section{Conclusion}
We examined how frontier LLMs adapt 60 English math word problems into 7 languages through entity-level annotation of 6,489 cultural transformations. Models agree on transformation type in only 62.5\% of cases and on specific substitutions in just 33.5\%, meaning that model choice is a cultural decision, not merely a technical one. All 21 language-model combinations show entropy collapse. Models prioritize surface markers while preserving structural features that embed culturally specific assumptions, and outputs that appear locally appropriate can still reflect cross-regional misattribution and cross-cultural contamination.

These results do not mean LLMs should be avoided in education. They offer real support for curriculum adaptation at scale, especially in under-resourced settings where teachers have limited time for manual adaptation. But our findings make clear that these outputs require active cultural review, not just as a precaution, but as a necessary step in the process. Does a substitution reflect the students' community, or does it reproduce a stereotype? Does a practice actually belong in the cultural context being invoked? Would the adapted problem feel meaningful, or would it feel subtly alienating to the students who encounter it? These are questions teachers already ask when adapting curriculum manually. AI assistance does not remove this series of questioning. If anything, it creates a familiar automation irony: adaptation becomes easier to generate, while oversight becomes harder because outputs often look appropriate enough that further scrutiny feels unnecessary \citep{bainbridge1983ironies}. That surface plausibility is not a feature. It is the problem. Our contribution is to show that the systems educators increasingly rely on often hide instability, cultural compression, and shallow substitution beneath a surface of local appropriateness. That surface plausibility is precisely what makes careful review necessary.

This study is the first stage of a broader research program. Ongoing work will deploy a human evaluation survey with native speakers across all seven languages, assessing how real learners and educators perceive these adaptations along dimensions of linguistic quality, cultural appropriateness, and reasoning preservation. A subsequent benchmarking study will ask where automated NLP metrics align with and diverge from the judgments of people with genuine cultural knowledge of the target communities. Further out, the patterns we document raise a deeper question: not just what models substitute, but why, what internal representations drive a model to reach for one name, food, or practice over another, and whether those associations reflect anything like genuine cultural knowledge. This is important because the use of LLMs for personalized learning are already in classrooms but the evaluation frameworks to assess them are not.

\section*{Acknowledgments}
We would like to thank the study participants for their time. We thank Alice Patania, Protiva Sen, Aviral Chawla, and Robin Sharma for their assistance with interpreting and translating text into the target languages, and a special thanks to Robin Sharma for assistance with labeling. We are grateful to Rakesh Kumar for an early conversation that encouraged us to pursue low-resource languages. We thank Julia Witte Zimmerman, Tabia Tanzin Prama, and Calla Beauregard for their guidance on the literature review.
Our team acknowledges support from the Alfred P. Sloan Foundation (Grant \#G-2024-22498) and the National Science Foundation (Grant \#2242829).

\section*{Ethical Considerations Statement}
This study analyzes outputs from commercial language models and does not involve human subjects, so no IRB review was required. All translations were generated through standard API access to Claude Opus 4, GPT-4.1, and Gemini 2.5 Pro under each provider's terms of service. The English source problems were drawn from GSM8K \citep{cobbe2021training} and the culturally annotated extension by \citet{karim2025lost}, both released for research use. No personally identifiable information appears in the source problems or the generated translations; the names that appear in the corpus are model-generated substitutions, not references to real individuals. The annotation work was conducted by members of the research team rather than crowdworkers. Where annotators lacked proficiency in a target language, we consulted lab members with relevant linguistic and regional background; these consultations were uncompensated collaborative discussions rather than paid annotation labor. We will release the full entity-level dataset, including source problems, model outputs, alignments, action labels, and back-translations, to support replication and further analysis.
\section*{Researcher Positionality Statement}
The authors are researchers based at The University of Vermont working at the intersection of natural language processing, computational social science, and education. This positionality shapes the work in ways we want to name openly. Most directly, it shaped what we chose to study. Every language in our study carries history. Sicilian holds layers of Greek, Arab, Norman, and Spanish presence in its proverbs and rhythms, but is considered vulnerable. Sindhi belongs to one of the oldest continuous cultural traditions in South Asia, carrying centuries of Sufi poetry and oral history dating back to the Indus Valley civilization, but is also vulnerable. Punjabi, shaped by Sikh, Sufi, and folk traditions on both sides of Partition, remains under-resourced in NLP relative to the cultural weight it carries. Many languages share this pattern of deep cultural roots and thin institutional standing; we chose this set because it is the one our lab can speak to and compare with higher resource languages from the same regions. A different team would likely have chosen differently, and the framing of ``resource levels'' we use here is itself a perspective from inside NLP rather than from inside these language communities.

Our positionality both supports and limits the analysis. Native and heritage familiarity with Urdu, Sindhi, Hindi, and Punjabi helped annotators identify some regional misattributions that might be less visible to outside readers. For languages where the formal annotators lacked direct expertise, we consulted lab members familiar with the relevant language or regional context for ambiguous cases. At the same time, the coding process may still reflect indirect biases, especially in borderline decisions about whether an entity was preserved, localized, or generalized. Our framework also reflects a computational perspective, foregrounding entity transformations, localization, and entropy collapse, whereas educators and community members might prioritize classroom fit, tone, or whether students feel addressed. Our planned human evaluation study is intended to address this asymmetry.

\section*{Adverse Impact Statement}
This work could be misread in several ways. First, our findings could be used to argue against deploying LLMs for educational adaptation in low-resource languages because outputs are unreliable. We do not endorse that interpretation. Manual adaptation is difficult to scale, and many under-resourced educational contexts may benefit from AI assistance if it is used carefully. Our argument is for cultural review of model outputs, not for withholding tools from the communities that may need them most.

Second, the adaptation spectrum is descriptive, not prescriptive. We do not claim that names should be localized, occupations should be preserved, or any specific transformation pattern is pedagogically correct. The tables describe what current models do, not what good cultural adaptation should do.

Third, per-language and per-model results should not be used as procurement benchmarks. Our findings are based on 60 source problems, three models, one prompting setup, and mid-2025 model snapshots. They document tendencies under a controlled setup, not general model quality for deployment.

Finally, LLM-generated adaptation creates a familiar automation irony: it may make culturally adapted materials easier to produce while making human oversight harder, because reviewers must detect subtle errors in fluent, plausible outputs. This risk is especially important in educational settings, where surface-level localization can appear inclusive while still producing stereotypes, regional misattribution, or culturally incoherent scenarios.
\bibliography{aaai2026}

\appendix
\section{Appendix I: Entity Type Schema}
\label{app:entity_type_schema}
The final schema contains 76 fine-grained entity types grouped into 26 super-categories. The initial source-side inventory contained 66 fine-grained types. During the labeling process, 10 additional fine-grained entity types were added because some target-side values introduced through cultural adaptation did not map cleanly onto the original source-side schema. These additions allowed the annotation framework to capture culturally specific substitutions more accurately without forcing them into ill-fitting source-side categories (Table~\ref{tab:super_categories}).

\begin{table*}
\centering
\small
\begin{tabular}{p{0.24\textwidth} p{0.70\textwidth}}
\hline
\textbf{Super-Category} & \textbf{Fine-Grained Entity Types} \\
\hline
Personal Names & person \\
Social \& Occupational Roles & family\_member, social\_group, participant\_role, occupation\_role \\
Food \& Drink Items & food\_item, baked\_good, dessert, raw\_ingredient, beverage \\
Culinary Practices & meal\_type, cooking\_method \\
Cultural Events & event \\
Life Course Milestones & life\_event \\
Consumer Electronics & electronics \\
Industrial Equipment & manufacturing\_equipment \\
Leisure, Play \& Sport & toy\_or\_game\_item, sport\_or\_game, sport\_or\_game\_event, activity \\
Places, Spaces \& Built Environment & residence, workplace, house\_type, retail\_venue, entertainment\_venue, restaurant \\
Education \& Learning & learning\_institution, class\_type, exam\_type, question\_type, book\_type \\
Religious Practices \& Places & religious\_place, holiday\_or\_holiday\_event \\
Geography \& Natural Places & country, state, administrative\_region, geographic\_region, city, village, landmark, natural\_feature \\
Money \& Currency Systems & currency\_name, currency\_symbol, currency\_denomination \\
Measurement \& Quantification & measurement\_unit \\
Markets \& Commerce & market\_type, online\_marketplace \\
Clothing \& Accessories & apparel, jewelry, scent\_type \\
Household Objects \& Tools & appliance, household\_item, decorative\_item \\
Transportation & vehicle\_type, vehicle\_part \\
Commercial Products & product\_category \\
Arts, Performance \& Media & music\_item, dance\_style, performance\_type, performance\_context, music\_instrument, creative\_work, movie, newspaper, digital\_content \\
Language, Narrative \& Symbolic Knowledge & language \\
Mythical Characters & mythical\_character \\
Nature \& Living Beings & animal, plant\_or\_flower, season, animal\_feed \\
Institutions \& Governance & organization, governing\_bodies \\
Materials \& Resources & fuel, precious\_metal, construction\_material \\
\hline
\end{tabular}
\caption{Mapping of 76 fine-grained entity types to 26 super-categories.}
\label{tab:super_categories}
\end{table*}

\section{Appendix II: Detailed Results by Entity Category and Language}
\label{app:results}

\begin{table*}
  \centering
  \small
  \begin{tabular}{llrrrr}
    \hline
    \textbf{entity\_type} & \textbf{super\_category} & \textbf{N} & \textbf{preserved} & \textbf{localized} & \textbf{adaptation balance} \\
    \hline
    vehicle\_part & Transportation & 21 & 1.000 & 0.000 & -1.000 \\
    meal\_type & Culinary Practices & 21 & 1.000 & 0.000 & -1.000 \\
    plant\_or\_flower & Nature \& Living Beings & 21 & 1.000 & 0.000 & -1.000 \\
    participant\_role & Social \& Occupational Roles & 105 & 0.990 & 0.010 & -0.980 \\
    life\_event & Life Course Milestones & 105 & 0.933 & 0.038 & -0.895 \\
    season & Nature \& Living Beings & 21 & 0.762 & 0.095 & -0.667 \\
    question\_type & Education \& Learning & 63 & 0.778 & 0.222 & -0.556 \\
    electronics & Consumer Electronics & 189 & 0.720 & 0.164 & -0.556 \\
    \hline
  \end{tabular}
  \caption{\label{tab:adaptation_spectrum_preserved_full}
  Entity types at the fully preserved end of the cultural adaptation spectrum (\textit{adaptation balance} \(< -0.5\)), ranked by adaptation balance score.}
\end{table*}

\begin{figure*}
\centering
\includegraphics[width=0.9\textwidth]{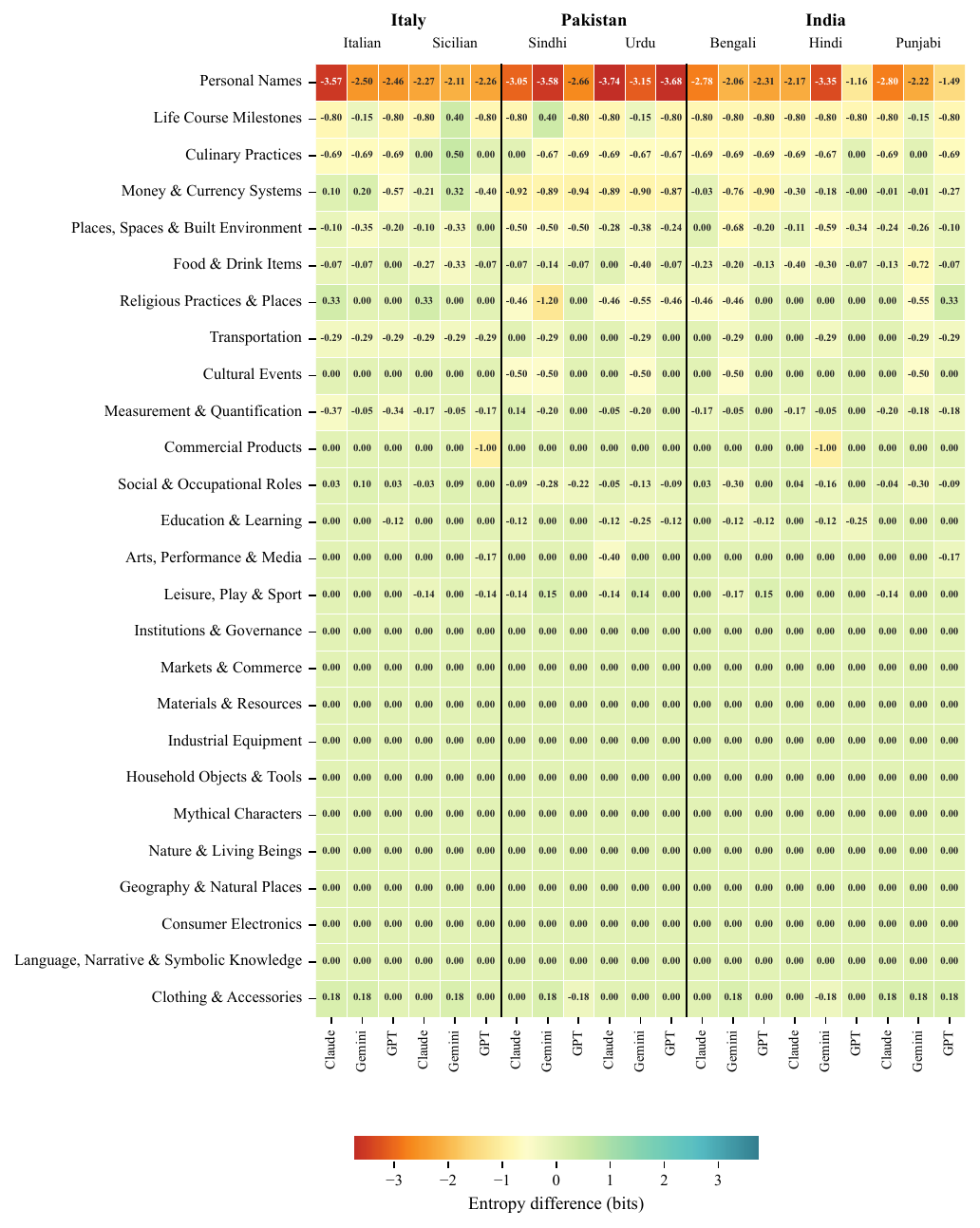}
\caption{Entropy collapse across entity categories, languages, and models. Cells show entropy difference in bits between English source entity values and back-translated model-generated values. Negative values indicate entropy collapse, where model outputs draw from a narrower set of entity values than the English source. Positive values indicate entropy maintenance or expansion. Personal Names show strong negative values almost everywhere, explaining their dominance in the overall pattern. Other categories vary substantially: Money \& Currency Systems shows strong collapse in Pakistani settings but weaker or slightly positive values in Italian and Indian settings.}
\label{fig:entropy_full_heatmap}
\end{figure*}

\begin{table*}
  \centering
  \small
  \begin{tabular}{llrrrr}
    \hline
    \textbf{entity\_type} & \textbf{super\_category} & \textbf{N} & \textbf{preserved} & \textbf{localized} & \textbf{adaptation balance} \\
    \hline
    product\_category & Commercial Products & 42 & 0.286 & 0.000 & -0.286 \\
    book\_type & Education \& Learning & 147 & 0.619 & 0.333 & -0.286 \\
    occupation\_role & Social \& Occupational Roles & 210 & 0.524 & 0.295 & -0.229 \\
    appliance & Household Objects \& Tools & 21 & 0.190 & 0.048 & -0.142 \\
    activity & Leisure, Play \& Sport & 84 & 0.452 & 0.321 & -0.131 \\
    retail\_venue & Places, Spaces \& Built Environment & 210 & 0.419 & 0.290 & -0.129 \\
    organization & Institutions \& Governance & 42 & 0.286 & 0.190 & -0.096 \\
    class\_type & Education \& Learning & 42 & 0.357 & 0.286 & -0.071 \\
    measurement\_unit & Measurement \& Quantification & 336 & 0.446 & 0.384 & -0.062 \\
    dessert & Food \& Drink Items & 21 & 0.524 & 0.476 & -0.048 \\
    sport\_or\_game & Leisure, Play \& Sport & 84 & 0.464 & 0.417 & -0.047 \\
    natural\_feature & Geography \& Natural Places & 63 & 0.302 & 0.270 & -0.032 \\
    mythical\_character & Mythical Characters & 42 & 0.143 & 0.143 & 0.000 \\
    cooking\_method & Culinary Practices & 63 & 0.000 & 0.000 & 0.000 \\
    workplace & Places, Spaces \& Built Environment & 63 & 0.397 & 0.413 & 0.016 \\
    animal & Nature \& Living Beings & 63 & 0.381 & 0.460 & 0.079 \\
    performance\_type & Arts, Performance \& Media & 21 & 0.381 & 0.476 & 0.095 \\
    animal\_feed & Nature \& Living Beings & 21 & 0.238 & 0.333 & 0.095 \\
    beverage & Food \& Drink Items & 105 & 0.410 & 0.524 & 0.114 \\
    toy\_or\_game\_item & Leisure, Play \& Sport & 105 & 0.371 & 0.486 & 0.115 \\
    event & Cultural Events & 84 & 0.238 & 0.369 & 0.131 \\
    baked\_good & Food \& Drink Items & 189 & 0.058 & 0.190 & 0.132 \\
    construction\_material & Materials \& Resources & 21 & 0.429 & 0.571 & 0.142 \\
    entertainment\_venue & Places, Spaces \& Built Environment & 42 & 0.405 & 0.548 & 0.143 \\
    learning\_institution & Education \& Learning & 42 & 0.381 & 0.548 & 0.167 \\
    jewelry & Clothing \& Accessories & 21 & 0.238 & 0.429 & 0.191 \\
    country & Geography \& Natural Places & 21 & 0.000 & 0.238 & 0.238 \\
    scent\_type & Clothing \& Accessories & 42 & 0.167 & 0.405 & 0.238 \\
    apparel & Clothing \& Accessories & 168 & 0.333 & 0.583 & 0.250 \\
    restaurant & Places, Spaces \& Built Environment & 21 & 0.000 & 0.286 & 0.286 \\
    currency\_symbol & Money \& Currency Systems & 651 & 0.005 & 0.292 & 0.287 \\
    \hline
  \end{tabular}
  \caption{\label{tab:adaptation_spectrum_contested_full}
  Entity types at the fully contested middle end of the cultural adaptation spectrum (\(\left|\textit{adaptation balance}\right| < 0.3\)), ranked by adaptation balance score.}

\end{table*}
\begin{table*}
  \centering
  \small
  \begin{tabular}{llrrrr}
    \hline
    \textbf{entity\_type} & \textbf{super\_category} & \textbf{N} & \textbf{preserved} & \textbf{localized} & \textbf{balance} \\
    \hline
    performance\_context & Arts, Performance \& Media & 21 & 0.000 & 1.000 & 1.000 \\
    newspaper & Arts, Performance \& Media & 42 & 0.000 & 1.000 & 1.000 \\
    music\_item & Arts, Performance \& Media & 42 & 0.000 & 0.952 & 0.952 \\
    currency\_denomination & Money \& Currency Systems & 21 & 0.000 & 0.952 & 0.952 \\
    currency\_name & Money \& Currency Systems & 105 & 0.000 & 0.943 & 0.943 \\
    person & Personal Names & 1176 & 0.041 & 0.957 & 0.916 \\
    market\_type & Markets \& Commerce & 42 & 0.071 & 0.929 & 0.858 \\
    religious\_place & Religious Practices \& Places & 21 & 0.095 & 0.905 & 0.810 \\
    food\_item & Food \& Drink Items & 126 & 0.079 & 0.865 & 0.786 \\
    residence & Places, Spaces \& Built Environment & 21 & 0.095 & 0.810 & 0.715 \\
    sport\_or\_game\_event & Leisure, Play \& Sport & 21 & 0.095 & 0.810 & 0.715 \\
    movie & Arts, Performance \& Media & 21 & 0.000 & 0.667 & 0.667 \\
    language & Language, Narrative \& Symbolic Knowledge & 63 & 0.175 & 0.825 & 0.650 \\
    online\_marketplace & Markets \& Commerce & 42 & 0.143 & 0.786 & 0.643 \\
    governing\_bodies & Institutions \& Governance & 21 & 0.000 & 0.619 & 0.619 \\
    dance\_style & Arts, Performance \& Media & 84 & 0.071 & 0.631 & 0.560 \\
    \hline
  \end{tabular}
  \caption{\label{tab:adaptation_spectrum_localized_full}
    Entity types at the fully localized end of the cultural adaptation spectrum (\textit{adaptation balance} \(> 0.5\)), ranked by adaptation balance score.}
\end{table*}

\begin{figure*}
\centering
\includegraphics[width=\textwidth]{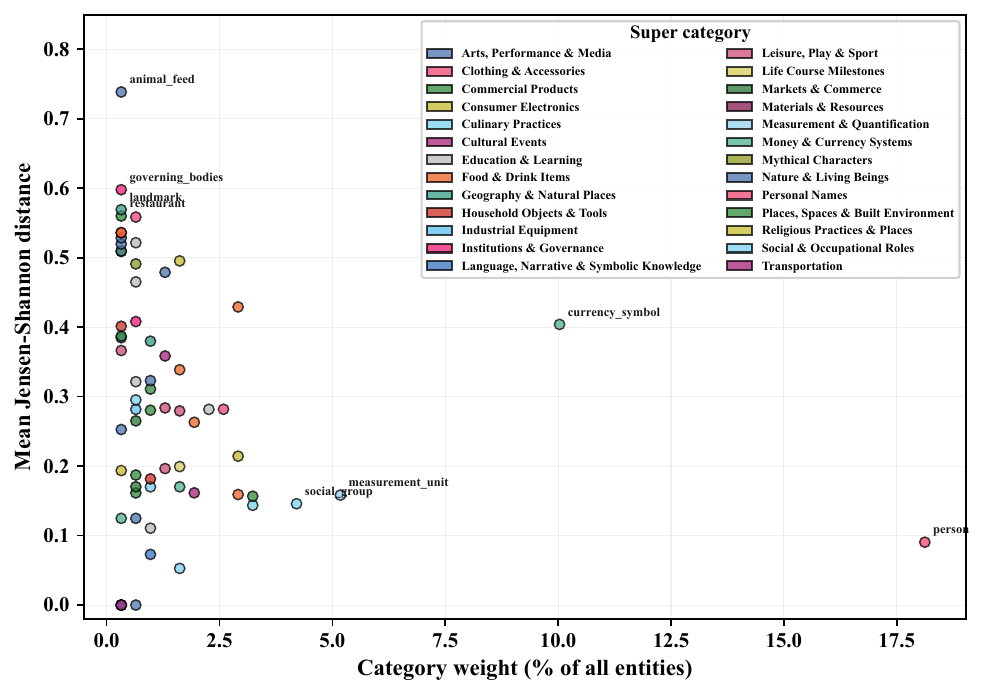}
\caption{Contribution of each entity supercategory to overall Jensen-Shannon divergence between languages. Each point represents a specific entity type (e.g., person, food\_item, currency\_symbol), colored by its broader super-category. The x-axis shows frequency in the dataset; the y-axis shows mean Jensen-Shannon distance across languages, measuring whether models take similar actions (preserved, localized, generalized, type changed, missing) for that entity type across the seven languages. Low JSD means languages get similar action distributions; high JSD means different languages get different actions. Personal names show consistent actions across languages, while categories like animal feed and governing bodies show inconsistent treatment.}
\end{figure*}

\begin{table*}[t]
  \small
  \centering
  \begin{tabular}{llrrr}
    \hline
    \textbf{Language} & \textbf{Model} & \textbf{$\Delta H$} & \textbf{$\Delta H_{MM}$} & \textbf{95\% CI} \\
    \hline
    Hindi & Gemini & -0.296 & -0.368 & [-0.581, -0.094] \\
    Urdu & Claude & -0.293 & -0.348 & [-0.630, -0.081] \\
    Sindhi & Gemini & -0.289 & -0.347 & [-0.617, -0.064] \\
    Urdu & Gemini & -0.286 & -0.349 & [-0.559, -0.100] \\
    Urdu & GPT & -0.269 & -0.320 & [-0.585, -0.061] \\
    Bengali & Gemini & -0.265 & -0.325 & [-0.459, -0.111] \\
    Sindhi & Claude & -0.250 & -0.301 & [-0.516, -0.065] \\
    Sindhi & GPT & -0.233 & -0.275 & [-0.486, -0.058] \\
    Italian & GPT & -0.209 & -0.253 & [-0.446, -0.055] \\
    Sicilian & GPT & -0.204 & -0.252 & [-0.407, -0.057] \\
    Italian & Claude & -0.201 & -0.239 & [-0.551, -0.003] \\
    Bengali & Claude & -0.197 & -0.241 & [-0.457, -0.035] \\
    Bengali & GPT & -0.192 & -0.227 & [-0.417, -0.038] \\
    Punjabi & Gemini & -0.192 & -0.240 & [-0.399, -0.048] \\
    Punjabi & Claude & -0.187 & -0.228 & [-0.447, -0.025] \\
    Hindi & Claude & -0.177 & -0.215 & [-0.380, -0.038] \\
    Sicilian & Claude & -0.151 & -0.182 & [-0.342, -0.017] \\
    Punjabi & GPT & -0.140 & -0.170 & [-0.293, -0.021] \\
    Italian & Gemini & -0.139 & -0.173 & [-0.356, \phantom{-}0.003] \\
    Hindi & GPT & -0.101 & -0.127 & [-0.224, -0.013] \\
    Sicilian & Gemini & -0.062 & -0.076 & [-0.258, \phantom{-}0.077] \\
    \hline
  \end{tabular}
  \caption{\label{tab:entropy_ci}
  Mean entropy difference by language and model, sorted from largest to smallest reduction. $\Delta H$ is the plug-in estimate; $\Delta H_{MM}$ applies the Miller--Madow bias correction. Confidence intervals are bootstrap 95\% intervals over the 26 super-categories in each combination (2,000 resamples). All 21 combinations are negative and 19 have intervals excluding zero; the exceptions are Italian and Sicilian under Gemini, the two smallest reductions. Corrected estimates are more negative than plug-in estimates in every combination.}
\end{table*}
\end{document}